\documentclass[letterpaper, 10 pt, conference]{ieeeconf}  
\usepackage{cite}
\usepackage{graphicx}
\graphicspath{ {images/} }
\usepackage{url}
\usepackage{booktabs}
\usepackage{adjustbox}
\usepackage{url}
\usepackage{courier}
\usepackage{verbatim}
\usepackage{mathrsfs}
\usepackage{dblfloatfix}
\usepackage{hyperref}
\usepackage{blindtext}
\usepackage{dirtytalk}
\usepackage{flushend}

\IEEEoverridecommandlockouts                              

\title{\LARGE \bf
MobiAxis: An Embodied Learning Task for Teaching Multiplication with a Social Robot*
}

\author{Karen Tatarian $^{1,4}$, Sebastian Wallk\"otter $^{2}$, Sera Buyukgoz$^{1,4}$, Rebecca Stower $^{3}$, and Mohamed Chetouani$^{4}$
\thanks{*This work is supported by the \href{http://www.animatas.eu/}{ANIMATAS} project }
\thanks{$^{1}$SoftBank Robotics Europe, 75015 Paris, France 
        {\tt\small \{name.surname\}@softbankrobotics.com}}%
\thanks{$^{2}$Uppsala University, Uppsala, Sweden 
        {\tt\small sebastian.wallkotter@it.uu.se}}%
\thanks{$^{3}$Jacobs University, Bremen, Germany 
        {\tt\small r.stower@jacobs-university.de}}%
\thanks{$^{4}$Sorbonne Universite, Institude for Intelligent Systems and robotics, CNRS UMR 7222, 75005 Paris, France 
        {\tt\small \{name.surname\}@sorbonne-universite.fr}}%
}

\begin{document}

\maketitle
\thispagestyle{empty}
\pagestyle{empty}

\begin{abstract}
The use of robots in educational settings is growing increasingly popular. Yet, many of the learning tasks involving social robots do not take full advantage of their physical embodiment. MobiAxis is a proposed learning task which uses the physical capabilities of a Pepper robot to teach the concepts of positive and negative multiplication along a number line. The robot is embodied with a number of multi-modal socially intelligent features and behaviours which are designed to enhance learning. This paper is a position paper describing the technical and theoretical implementation of the task, as well as proposed directions for future studies. 
\end{abstract}

\vspace{0.3cm}

\section*{INTRODUCTION}

{}

Social robots can offer many advantages over and above pure computer or tablet based learning activities, such as physical navigation of the environment, multi-modal social behaviours, and non-verbal communication \cite{Fridin2014, Leyzberg2012, Kennedy2015a}. However, many learning tasks which involve the use of social robots fail to take full advantage of the physical and navigational capacities of the robot. As such, this paper proposes a model which involves an educational learning task (MobiAxis) between a child and Pepper robot. MobiAxis is designed around navigation, specifically targeting the concepts of number lines and multiplication of positive and negative numbers. This task was chosen as it allows for physical manipulation and navigation in space, maximizing the use of the robots physical capabilities. In addition, learning of mathematics has shown to benefit from the manipulation of tangible elements \cite{Tsang2015}.  MobiAxis is also designed to investigate how specific socially intelligent behaviors can most benefit children's learning outcomes.

The task is comprised of four phases: a demonstration phase (where the robot performs an example of the task and the child observes), a supervision phase (robot acts as tutor whilst the child performs the task), a teaching and learning phase (child gives robot instructions) and cooperation phase (child and robot work together). Throughout these phases, the robot implements a number of multi-modal socially intelligent behaviors, such as:
\begin{itemize}
    \item Physical navigation of the environment.
    \item Proactive engagement detection and adaptation.
    \item Turn-taking. 
    \item Non-verbal communication through physical gestures.
    \item Personalized feedback.
    \item Adaption of different pedagogical roles (peer, tutor, novice).
\end{itemize}

One of the goals of MobiAxis is to explore how these behaviours influence children's learning outcomes. How these behaviours can be used to moderate social elements of the interaction such as trust, perceived agency, and engagement are also further research questions of interest. 


\section*{BACKGROUND}
The development of numerical knowledge during childhood is important in our current culture; for example, mathematical ability at age 7 can predict later socio-economic status (SES) almost as well as birth SES does \cite{Ritchie2013}. Further, there is evidence that early math ability is not only strongly linked to later math achievement, but also influences later reading achievement \cite{Duncan2007}. In light of this significance, we chose arithmetic as the topic for this learning task.

Furthermore, research in psychology has shown that the brain regions used for numerical computations, and in extension the mental number line, are the same regions we use for spatial thinking and vision \cite{Dunlosky2019}. In line with these findings, as well as the large body of evidence suggesting that physical manipulation helps learning \cite{Manches2010, Vitale2014, Abtahi2016, Tsang2015}, we propose to use a robot moving along a physical number line to show the process of multiplication. A similar task was used by Tsang et al. \cite{Tsang2015}; however instead of including a social robot, they layer a second task (called cancellation model) on top, to test if this will lead to greater learning. 




From the human-robot interaction (HRI) perspective, there has already been great interest in the use of robots for education \cite{Spolaor2017, Pachidis2019}, and math education in particular \cite{Zhong2018, Leoste2019}. 
The NAO\footnote{NAO robot by SoftBank Robotics \url{www.softbankrobotics.com/emea/en/nao}} robot has been used to teach geometric thinking \cite{Pinto2016,Keren2014}, and there is evidence that robotic kits, e.g., LEGO robots, are beneficial for math learning \cite{Julia2016a,Padayachee2016,Chang2012}.

However, how the specific behavior of the robot can influence these learning outcomes is an open question. Whilst there is some evidence to suggest multi-modal social behaviours in robots can increase learning gains \cite{Belpaeme2018, kennedy_tony}, other findings question these conclusions \cite{Kennedy2015b}. This research therefore also aims at determining the role of social behaviours, particularly in the domain of math education.

Another aspect that makes our task unique is its kinesthetic element, which is seeing a growing amount of supporting evidence \cite{macedonia2017your, kim2011children, carbonneau2013meta}. Rather than being seated or standing in one place during the interaction - as is the case in classic, lecture-style learning, as well as most robotic studies on the topic - our task is based on movement. Both the robot and the child move along the number line, which is distinctive to physically situated robots. We hypothesize that this added element of physical manipulation will be advantageous to learning outcomes.

\section*{MobiAxis: Set-Up}
MobiAxis is an educational task that requires a printed physical axis laid on the floor\footnote{full scale image attached in supplementaries} as a carpet, and an embodied robot.  Both the robot and child navigate along the axis. As such, the task maximizes use of the physical space to allow for better visualization and understanding of the concepts presented in each level. MobiAxis is aimed at teaching students between the ages of $10$ and $12$ the mathematical concepts related to multiplication of positive and negative numbers (i.e., how two negative numbers when multiplied together make a positive). The task is based around the commonly used format of number lines, but expanded to include orientation and navigation along an axis (promoting understanding of the concepts of direction and magnitude). Each level is made of three phases and a bonus phase (see \textit{\nameref{sect: phases}}). \\

\subsection*{Key Mathematical Concepts in Learning Task}
At this point it is important to define the key words: orientation and direction. Orientation refers to the starting position of the navigator (robot or child). From zero, facing towards the positive axis constitutes a positive orientation, whereas facing towards the negative axis constitutes a negative orientation. The orientation is determined by the sign of the first number in the multiplication equation. The direction of movement along the axis can then be either forwards or backwards according to the current orientation. The direction of movement is determined by the sign of the second number in the multiplication equation. For example, $+2 \times -3$, the first number is $+2$ implying the orientation is positive (facing towards the positive axis) and the second number is $-3$ implying movement in the backwards direction along the axis. Note that if the example were reversed to be $-2 \times +3$ the orientation would be negative and the direction forward, leading to the same final position. 

\subsection*{Materials \& Methods} \label{subsection: set up}
\begin{figure}[htb]
    \centering
    \includegraphics[width=3in]{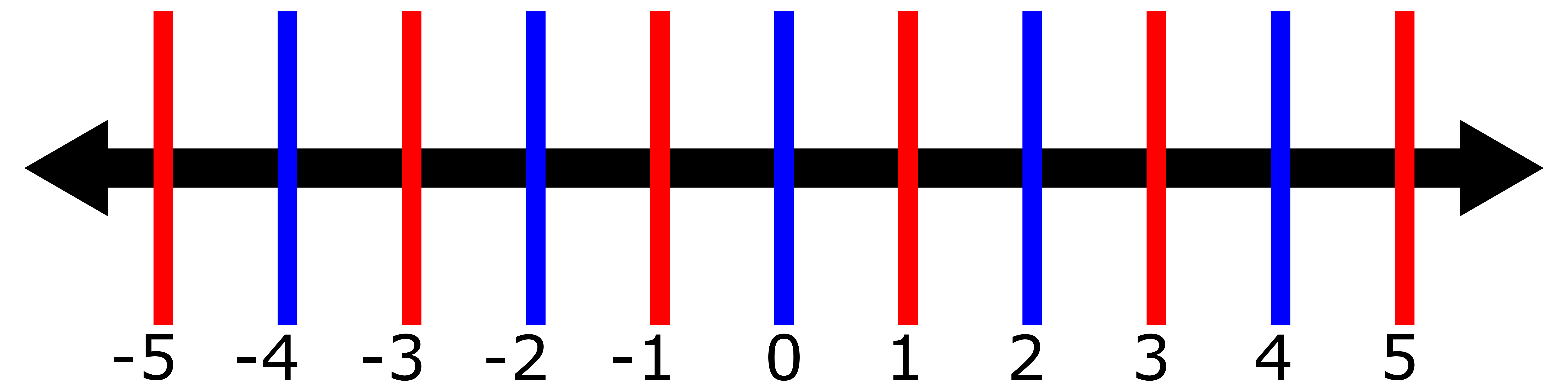}
    \caption{A shorter version of the axis used in the Learning Task, MobiAxis}
    \label{fig:axis_five}
\end{figure}

The axis used in MobiAxis is made up of red and blue bars, which alternate between odd and even numbers, respectively. The robot’s eyes also change colour to coordinate with the bar(unit) it stops on. This allows the child to detect accurately where the robot has stopped. In addition, the axis has intervals of step units of $26~\textrm{cm}$ and the axis goes from $-10$ to $+10$. Figure \ref{fig:axis_five} shows a shorter version of the axis going from $-5$ to $+5$.\\

\begin{figure}[htb]
    \centering
    \includegraphics[width=3.5in]{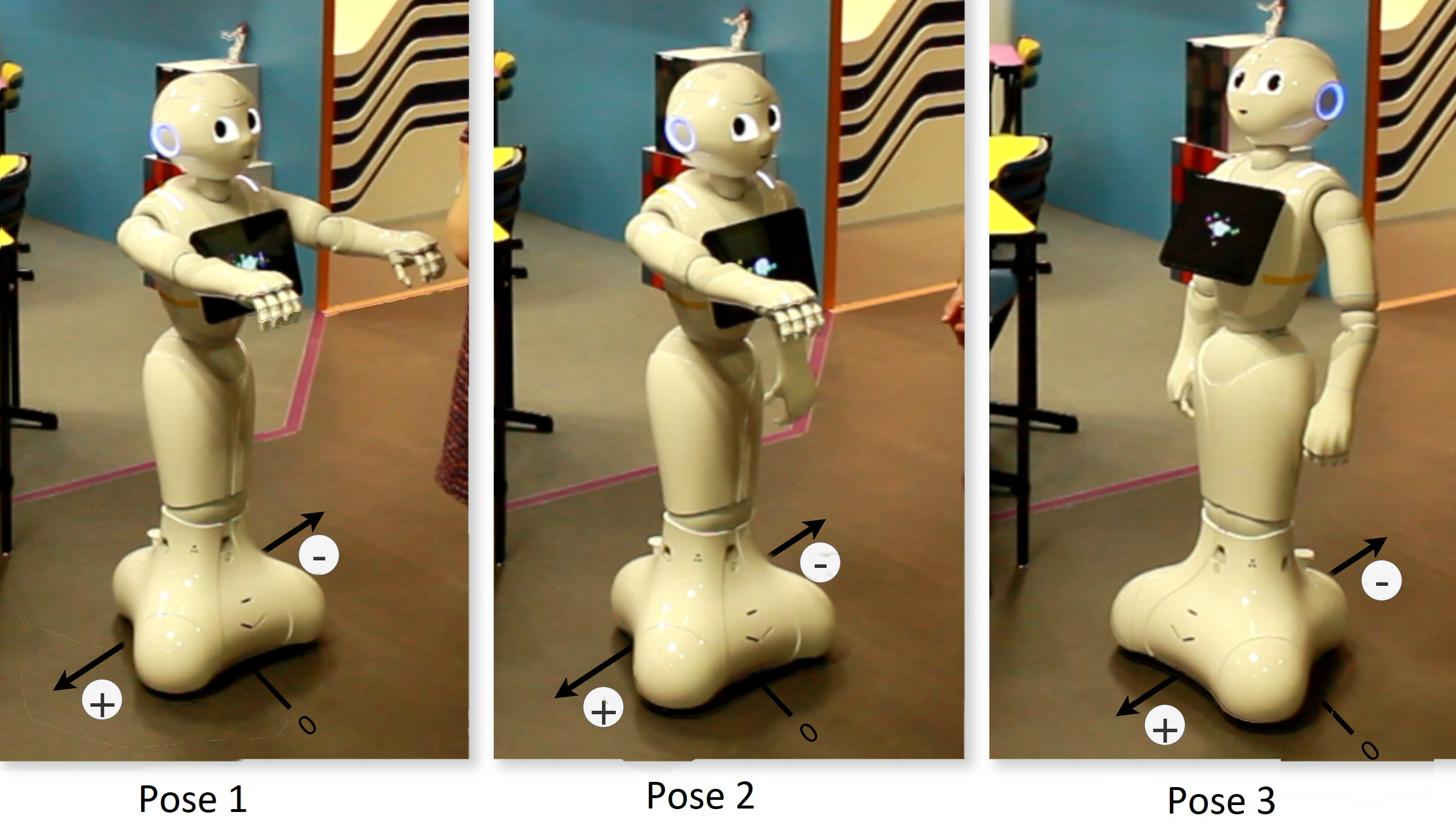}
    \caption{The three poses of the Robot for selecting orientation \& direction}
    \label{fig:poses}
\end{figure}

The robot used for MobiAxis is Pepper \footnote{Pepper robot by SoftBank Robotics \url{www.softbankrobotics.com/emea/en/pepper}}. The robot has three main poses, which are shown in Figure \ref{fig:poses}.  At the beginning of the task, it is located at \textit{pose1}, where it is positioned at the zero position of the axis, facing the child, with both of its arms lifted $90$ degrees. First, the orientation of the robot is chosen by touching the hand of the side that the robot needs to face (right hand for positive orientation, left hand for negative orientation). This causes the robot to move to \textit{pose2}, where it is still at the zero position of the axis and facing the child but this time only the chosen arm for orientation is raised. Second, the number of times the selected hand is tapped specifies the step size, where $step size = number of taps \times [fixed step unit]$ with the step unit being prior set and constant throughout all levels. Third, the direction is selected on the tablet of the robot. At this point, the robot switches to \textit{pose3}, where it is still at the zero position but facing the chosen orientation, having both arms down. Finally, the number of steps, which is how many times the step (which has the length of the registered step size) is repeated, is registered by the number of times the robot's head is tapped. In Figure \ref{fig:poses}, the \textit{pose2} shows the right hand raised indicates that the positive orientation was selected while \textit{pose3} has the robot turned facing the positive side showing that the robot has oriented positively. Note, in \textit{pose3} of Figure \ref{fig:poses} only shows orientation, if the direction was selected to be positive then the robot would move forward on the axis and if selected negative it would move backwards on the axis.\\

\subsection*{Implementation on the Pepper robot}
The Pepper robot runs on the operating system NAOqi \cite{amit}. NAOqi  has an application programming interface (API), which is a set of different modules for controlling and accessing the robot, and creating applications. In our application, the NAOqi APIs\footnote{NAOqi Documentation \url{http://doc.aldebaran.com/2-5/naoqi/index.html}} are used for low-level robot control and perception. The Python software development kit (SDK) of NAOqi is used for the programming task itself. In this implementation, we used version 2.5 of NAOqi and version 1.8 of the Pepper robot. One major improvement in this version, compared to the previous versions, is an updated localization framework. It improves dead reckoning for both, transnational and rotational movements, providing good enough out of the box accuracy for navigation in MobiAxis.

\begin{figure}[htb]
    \centering
    \includegraphics[width=3.5in]{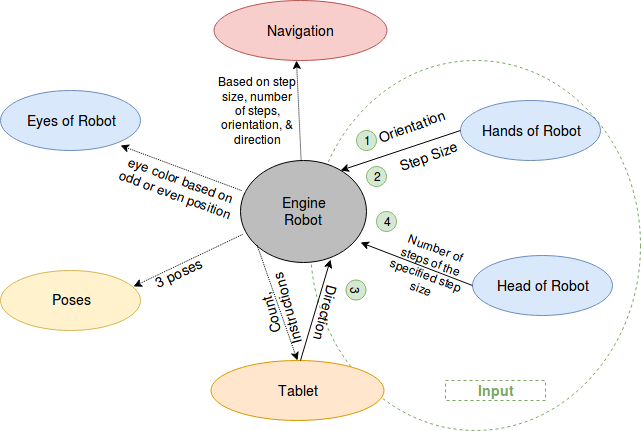}
    \caption{Model of Implementation}
    \label{fig:implementation_draft}
\end{figure} 

\begin{figure}[ht]
    \centering
    \includegraphics[width=3.5in]{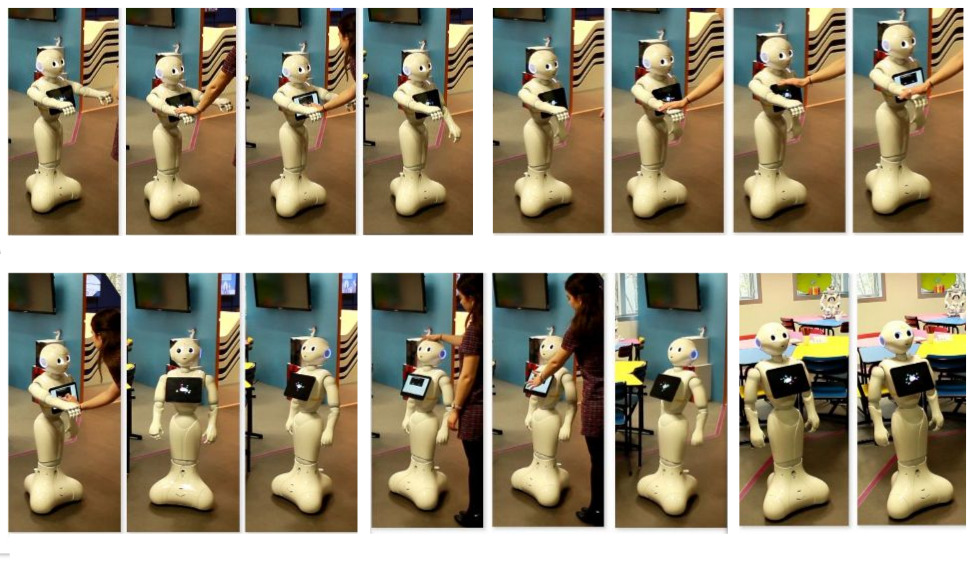}
    \caption{Time-line for instructing the robot to complete the mathematical equation}
    \label{fig:tasktimeline}
\end{figure}

As shown in Figure \ref{fig:implementation_draft}, we have four different types of inputs, which are received from different services of NAOqi APIs. Touch sensors, which are placed in the hands and head of robot, are accessed through the ‘ALSensors’ services. Through that service, the listener of each sensor is raised when its respective sensor is touched. During the task,  we only subscribe to the sensors when it is necessary. For example, after the desired hand of the robot has been selected, we unsubscribe from the sensor of the arm that was not chosen, and instead, the script moves into the counting loop for calculating the step size based on the sensed taps on the chosen arm. Meanwhile, the tablet is displaying the registered choices made by the user. After this step, the user is asked to select the direction on the tablet. 'ALTabletService' services are used to display a certain output on the tablet, in addition to receiving input from it. On the other hand, the 'ALMotion' service is used to control the movement of the robot. It is used to control the robot's transition from one pose to another,  its orientation, and its navigation in the selected direction. The animations of poses are defined as a result of angle interpolation done by setting the position of each key frame and the time interval in which the action appears. Figure \ref{fig:poses} shows the poses and the transition between them. Furthermore, to match the reached destination on the axis on which the robot is navigating 'ALLeds' service is used to specify the eye colour, which is either red if the number reached is odd or blue if it is even. 
Finally, 'ALTextToSpeech' service is utilized to create the speech of the robot while Pepper is giving the instructions, counting the step size, and asking for verification from the user. 
Figure \ref{fig:tasktimeline} shows how task iterates each actions by employing the services described above.

\begin{table*}[ht]
\centering
\caption{Summary of Phases in Each Level of MobiAxis}
\label{tab:topics1}
\begin{adjustbox}{width=\textwidth}
\small
\begin{tabular}{@{}llp{60mm}p{60mm}@{}}
\toprule
\textbf{Phase} & \textbf{Role of Robot}         & \textbf{Description}            &\textbf{HRI Interest}         \\ \midrule
\texttt{Phase 1:}Demonstration        &Demonstrator    &Robot shows child an example by performing it   & Basis of trust model, \& role of robot                                                                 \\
\texttt{Phase 2:}Supervision        & Tutor    &Robot supervises the child as they complete an exercise on the axis as described in subsection \nameref{subsection: set up}        & Demonstration of turn-taking, basis for trust model, \& role of robot                                        \\
\texttt{Phase 3:}Teaching \& Learning       & Peer    &Child instructs the robot on how to move while robot gives feedback        & Model of trust, theory of mind, how the robot learns from human input, learning by teaching, feedback system, role of robot \& displaying multi-modality of social intelligence           \\
\texttt{Phase 4/Bonus Phase:}Collaboration        & Peer Collaborator    &Child and robot both work together towards solving a problem        & Model of trust, displaying multi-modality of social intelligence, role of robot, degree of collaboration, displaying turn-taking, \& pro-activity           \\\midrule
\end{tabular}
\end{adjustbox}
\end{table*}

\section*{Phases of the Learning Task}
\label{sect: phases}
 MobiAxis is comprised of several mathematical lessons related to multiplication, direction, and orientation. Each lesson is covered in one level. For instance, \textit{level1} covers multiplying positive numbers, while \textit{level2} and \textit{level3} go over multiplying positive and negative numbers and multiplying arbitrary numbers, respectively. Each level is made up of 3 phases, plus a bonus phase. In each phase, the robot plays a different pedagogical role to explore additional HRI research questions. The phases are summarized in Table \ref{tab:topics1}. Between each phase, the robot and the child interact to discuss the task. \\
 \vspace{0.15cm}
The phases of each level are designed to increase the learning gains from the activity and inspired by the building blocks presented in a blog article by Concordia University-Portland \footnote{the online blog article written by Concordia University - Portland  \url{https://education.cu-portland.edu/blog/classroom-resources/basic-math-teaching-strategies/}}. The teaching strategies for teaching mathematics include repetition, time testing, pair work, and manipulation tools.\\

First, repeating and reviewing previous formulae help facilitate learning and memorizing \cite{repetition}. As such, MobiAxis repeats the same lesson in the several phases, but in each phase with some variation. Second, time testing of the material is used to keep track of the student\'s progress. In MobiAxis, time testing takes place in the interactions between the robot and student in between the phases. Third,  students can ameliorate their critical thinking and problem solving skills, 
in addition to expressing themselves by group work in mathematics \cite{pair_work}. For this reason, MobiAxis includes pair work between robot and child through the last two phases of each level. Finally, the manipulation tools used to teach mathematics are implemented through the navigation on the axis and the use of tablet on the robot.\\
The first phase is \textit{Demonstration}, where the child learns by observing,
followed by the second phase \textit{Supervision}, where the aim is for the student to learn by doing the task alone for the first time and receiving feedback from robot about their decisions. This is then followed by the third phase \textit{Teaching \& Learning}, where the student is able to learn by teaching the robot,
and finally for advanced lessons there is a fourth phase \textit{Collaboration}, where the students collaborates with the robot to solve a problem. The sequence of the phases is chosen so that the child can experience several learning strategies and we can track the learning gains throughout. \\


\subsubsection*{Phase 1: Demonstration}
This is the first phase of a level where the robot demonstrates to the child what is orientation, step size, step numbers, and direction by performing an example. For instance, a possible scenario would be the one shown in table \ref{tab:scenario1}. The purpose of the first phase is to have the child learn by observing before attempting to do it himself. It is the first part of the interaction between the child and the robot, and as such forms a basis for the trust model between them. In addition, the child is exposed to the first role of the robot as a demonstrator.


\begin{table}[htb!]
\begin{flushleft}
\caption{Phase 1: Demonstration, an example scenario}
\normalsize
\label{tab:scenario1}
\end{flushleft}

\begin{adjustbox}{width=0.45\textwidth}
\begin{tabular}{|p{8cm}|}
\toprule
Pepper:\say{ To calculate $-2 \times 3$, I first need to work out which way I should turn, and the size of my steps. The first number is negative 2, which means I need to face the negative axis. This is also means my step size is 2 units}. \\
\textit{Pepper turns 90 degrees }\\
Pepper: \say{Now I need to calculate my direction and the number of steps I should take. The second number is positive three, so I need to move forward. So now I need to repeat the step size of 2 units 3 times}.\\
\textit{Pepper moves to $-6$}\\
Pepper: \say{When I multiply -$2 \times 3$, I get $-6$.}\\
\midrule
\end{tabular}
\end{adjustbox}
\end{table}
\subsubsection*{Phase 2: Supervision}
 In the second phase, the robot becomes an observer and the child performs the task alone by navigating along the axis. The child has to physically perform the orientation and direction and walking on the axis. The robot gives feedback to the child based on the performance and the decision he/she makes. Moreover, it proactively corrects the child’s decision. For example, the robot can warn the child if he/she is looking in the wrong direction and and/or calculating incorrectly. This phase allows us to explore the child's theory of mind regarding the robot and the trust model in the interaction. Furthermore, the feedback system allows us to test the right and wrong modalities and explore how accepting the child is of the suggestions and the comments made by the robot. The phase ends when the child completes the exercise. This forces a turn-taking interaction between the child and robot. Additionally, the pedagogical role of the robot changes to tutor. 

\vspace{0.3cm}

\subsubsection*{Phase 3: Teaching \& Learning}
%
In the third phase, the robot listens to the child for instructions on how to complete the exercise. The robot navigates along the axis whilst the child gives instructions. Table \ref{tab:scenario3} shows an example of scenario for Phase 3. In this phase, we can test the effect of the robot\'s proactivity on the child\'s engagement in the game. When the child loses focus, the robot can remind him or her to continue with the task. Finally, we can also study the way the robot learns from the human input. In a sense, the child is also teaching the robot how to solve the problem.

\begin{table}[htb!]
\caption{Phase 3: Teaching \& Learning, an example scenario}
\normalsize
\label{tab:scenario3}
\begin{adjustbox}{width=0.45\textwidth}
\begin{tabular}{|p{10cm}|}
\toprule
Pepper: \say{
Wow! You seem to have mastered this!
Can you help me to calculate $-2 \times 3$? First, tap one of my hands to tell me what side I should face.}\\
Child: \textit{taps left hand}\\
Pepper: \say{That is great! Please tap on my hand again to tell me how many units the size of my steps should be}.\\
Child: \textit{Taps left hand twice}
\textit{Pepper turns to face negative axis}\\
Pepper: \say{Which direction should I go now? Use my tablet to tell me forward or backward}.\\
Child: \textit{Taps ‘forward’ on the tablet}\\
Pepper:  \say{Tap my head to tell me how many times you would like me to repeat the unit step size of 2}\\
Child: \textit{Taps head 3 times}\\
Pepper moves forward along the negative axis to $-6$.\\
\bottomrule
\end{tabular}
\end{adjustbox}
\end{table}

\vspace{0.3cm}

\subsubsection*{Phase 4: Collaboration}
The fourth phase only exists for advanced levels. In this phase, the robot acts as a collaborator and the child and robot have to perform the task together to achieve the final answer. For example, the child can be the one who chooses the orientation, but it would the robot choosing the direction. In the future, the axis can also be used to learn how to add vectors and as such the child can be one vector and the robot another and they have to collaborate to find the answer. \\

\subsection*{Evaluation of Learning Task}
The learning task involves 3 main metrics by which to evaluate the effectiveness of the robots social behaviours; learning, engagement, and trust.

\subsubsection{Learning}
Learning in this context refers to the learning gain in the specific topic of interest being taught (in this case, multiplication of positive and negative numbers). Children will be administered a pre and post-test before and after interacting with the robot to assess their knowledge of multiplication. The scores on these tests can then be compared in order to detect if there is any improvement. Additionally, throughout the interaction, other metrics such as number of errors made by the child, time taken to complete an exercise, and advancement through the phases can be used to assess learning.

\subsubsection{Engagement}
Engagement can broadly be defined as the establishment and maintenance of connections between two (or more) agents involved in an interaction \cite{SIDNER2005140}. Engagement can further be broken down into two sub components; task engagement, which refers to engagement in solving a problem, and social engagement, which refers more to the social relationship being formed between participants \cite{Corrigan}. MobiAxis will include elements of both social and task engagement. Both objective and subjective (self-report) measures can be used, including, but not limited to, checking for length and frequency of eye contact with the robot, and self-report questionnaires. 

\subsubsection{Trust}
Similarly to engagement, trust can be broken down into two components; social and competency \cite{Gaudiello2016, VanStraten2018}. Social trust refers to more affective based trust with regards to the social relationship formed with the robot, whereas competency trust is an evaluation of the robots perceived capabilities. Competency based trust can be evaluated through willingness to adhere to the robots suggestions \cite{rebecca}. Social trust can be measured through self-report measures such as desire to be friends with the robot, or objective measures such as amount of information disclosed to the robot throughout the interaction \cite{rebecca}. 

\section*{DISCUSSION}
Pepper, in the current application, navigates and acts as a peer for the student, as well as a collaborator and teacher, switching roles depending on the situation, as suggested in Section \textit{\nameref{sect: phases}}. However, the robot can have further social features that can improve the interaction between it and the child. In the next stage of the application, we would like to add socially intelligent characteristics to better explore HRI research questions related to engagement, trust, and learning. First, we aim at designing an engagement detection model, which can be used during the interaction that takes  place in between each phase. This is used to help the robot be more proactive. For instance, if engagement is detected to be too low, the robot can act proactively to increase it or ask for teachers help. Thus, it is important for the robot to know the user's needs and adapt to them. In addition, it is important for the robot to exhibit turn-taking mechanisms to be able to play the role of a peer and collaborator. Multi-modal, socially intelligent behaviors are also necessary in order to to be perceived as an intelligent agent. In addition, we would like to add a feedback and reward system. The robot would give feedback to the student specifically in the 
second
and fourth phases to help guide him/her to the right answer while letting him/her learn and explore. The robot would reward the child with points for getting the right answer and improving learning gains. The feedback system would also adapt to the child's progress.\\
\vspace{0.2 cm}
The reason behind including such features and models is to be able to better explore how socially intelligent behaviors influence children's learning outcomes. Firstly, multi-modal social behaviors and their perception as a unified construct are highly important in order to attain desirable outcomes with social robots and better learning data \cite{kennedy_tony}. In addition, the findings in \cite{kennedy_tony} suggest that there is a strong positive correlation between ratings of tutor nonverbal immediacy and performance in a one-on-one maths tutoring task between robot and student. However, the exact effects that the verbal and non-verbal behaviours, such as emotional expression and multi-modal, socially intelligent behaviors (\textit{e.g.,} joint attention, gaze mechanisms, and gestures), have on children’s interactions with social robots is still unclear, especially for children’s trust in robots and the benefits for learning \cite{rebecca}.
Secondly, proactive behaviour is intended to not only take initiative when an action occurs but to also take initiative before an action happens by anticipating the needs of the user. The proactive model that we hope to implement would include anticipating user needs, improving the robot’s knowledge such as validation results or asking missing points \cite{c10}, increasing engagement and interaction by proactively seeking users for interaction \cite{c8}, learning algorithms, and considering user's actions and feelings such as cooperative manipulation \cite{c4, c3} and the balance for the robot not to become annoying \cite{claire}.  We hypothesize that the proactive behavior of the robot would increase the engagement of the user. It has been shown that such proactive behaviour may be vital to allow the robot to effectively engage users \cite{Liu2018}. 

\section*{FUTURE WORK}
In the real world, it is impossible to provide every single student a one-on-one teacher that can take the time to get to know the child and his/her learning style. However, robots may provide a solution to fill in the gaps. Robots like Pepper can navigate and manipulate the physical world just like the child. Whilst MobiAxis is still in its early stages, it aims at exploring social robotics and learning by using movement and the physical environment. The next steps of this learning task include moving the application to the 2.9 Pepper SDK for Android developed by SoftBank Robotics \footnote{Documentation can be found here: \url{https://qisdk.softbankrobotics.com/sdk/doc/pepper-sdk/index.html}} and conducting a pilot study. 

\section*{ACKNOWLEDGMENT}

This project has received funding from the European Union’s Horizon 2020 research and innovation programme under grant agreement No 765955. 

\bibliographystyle{plain}
\bibliography{PsychLit.bib,HRILit.bib,proactive.bib,introlit.bib}

\end{document}